\documentclass[runningheads]{llncs}
\usepackage{graphicx}
\usepackage{cite}
\usepackage{multirow}

\usepackage{bbding}
\begin{document}
\title{MedSim: A Novel Semantic Similarity Measure in Bio-medical Knowledge Graphs\thanks{Supported by the National Natural Science Foundation of China (No.61602013) and the Shenzhen Key Fundamental Research Projects (Grant No. JCYJ20170818091546869 and JCYJ20170412151008290).}}
\titlerunning{MedSim}
%
%
\author{Kai Lei\orcidID{0000-0001-9197-895X} \and
Kaiqi Yuan\orcidID{0000-0003-0985-3790} \and Qiang Zhang\orcidID{0000-0003-3922-2789} \and
Ying Shen\Envelope\orcidID{0000-0002-3220-904X}}
\authorrunning{K. Lei et al.}
%
\institute{School of Electronic and Computer Engineering,\\ 
Peking University Shenzhen Graduate School, Shenzhen, China\\
\email{\{leik,shenying\}@pkusz.edu.cn, kqyuan@pku.edu.cn, zhangqiang@sz.pku.edu.cn}}
\maketitle              
\begin{abstract}
We present MedSim, a novel semantic \textbf{SIM}ilarity method based on public well-established bio-\textbf{MED}ical knowledge graphs (KGs) and large-scale corpus, to study the therapeutic substitution of antibiotics. Besides hierarchy and corpus of KGs, MedSim further interprets medicine characteristics by constructing multi-dimensional medicine-specific feature vectors. Dataset of 528 antibiotic pairs scored by doctors is applied for evaluation and MedSim  has produced statistically significant improvement over other semantic similarity methods. Furthermore, some promising applications of MedSim in drug substitution and drug abuse prevention are presented in case study.

\end{abstract}
\section{Introduction}
Semantic similarity metric is widely used in medical information retrieval ~\cite{ref_article2} and medical knowledge reasoning. The most promising application scenario is therapeutic substitution, also known as therapeutic interchange and drug substitution. It is the practice of replacing one prescription with chemically different drugs that are expected to have the same clinical effect. Medicine semantic similarity measure plays an important role in this context by enabling a proper interpretation of drug information~\cite{ref_article3}.

Unlike conventional similarity measures, semantic similarity methods based on Knowledge Graph (KG) have been proven effective in nature language processing and information retrieval ~\cite{ref_article1}. KG is a type of graph structure that records massive entities and relations, such as FreeBase~\cite{ref_article4} and DrugBank~\cite{ref_article25}. DrugBank is a well-known bioinformatics KG for its broad scope and great integrity, which classifies medicines with multi-category bio-medical knowledge bases. 

The existing KG-based semantic similarity can be classified into structure-based similarity measures, and corpus-based ones. Aimed at heterogeneous networks, PathSim ~\cite{ref_article9}, and random walk ~\cite{ref_article11} are introduced to compute semantic similarity among the same type of objects by fully utilizing the path information. IC-based measures primarily rely on the contextual information of words, which usually measures the general semantic relevance between two words ~\cite{ref_article12}. Distributed representation methods ~\cite{ref_article14} calculate semantic similarity by transforming concepts into dense low-dimensional vectors learned from the large scale of corpus. Hybrid measures combine structural features with corpus features to overcome data sparseness and data noise. A weighted path method (Wpath) ~\cite{ref_article19} is proposed by employing both path length and information content. SimCat~\cite{ref_article20} incorporates category corpus and relationship structure information. However, due to the limited coverage of KGs, the aforementioned measures cannot be directly applied to a specific domain.

The bio-medical field has complex and diverse terminology, hierarchies and attributes to be considered. Pedersen et al.~\cite{ref_article21} presented a cluster-based approach with new features and evaluated this method for two different bioinformatic knowledge bases within the UMLS. Traverso et al. ~\cite{ref_article22} proposed GADES to compare entities in bioinformatic knowledge graphs by encoding the KG in aspects, e.g., hierarchies, neighborhoods, and specificity. Even though these semantic simlarity methods consider the characteristics in the bio-medical field, they often rely on a limited number of data sources and are validated in a limited scale of dataset. Besides, they exclusively depend on the KG-mined features rather than fully utilized the textual information of KG.

To address the problem above, we propose MedSim, a novel semantic similarity method based on public well-established medicine KG and multi-category data sources, to study the therapeutic substitution of antibiotics. Antibiotics are extensively applied as antimicrobial agents in disease treatment but the abuse of antibiotics is becoming increasingly serious. We consider the semantic similarity between two antibiotics in bio-medical category using not only medicine-specific features but also the structure and corpus information of DrugBank which is freely accessible. To our knowledge, the biomedical domain never witnesses the standard human rating datasets for semantic similarity publications. A dataset labeled by doctors which is much larger than other medical similarity methods~\cite{ref_article23,ref_article24} is applied to  evaluate MedSim. To make our method more reproducible, the dataset is freely accessible in Github \footnote{https://github.com/YuanKQ/MedSim-antibiotics-labeled-dataset
}. The main contributions of this work can be summarized as followed: 
\begin{itemize}
 \item To improve the interpretability of drug property and the context-based word representation, the one-dimensional vector of medicine-specific features is transformed to multi-dimensional weighted vectors. The medicine-specific features reflect medicine characteristics and can be extended to all types of medicine.
 \item To reduce the noises introduced by hierarchical structure on semantic similarity metrics, we employ a KG-based hierarchy embedding feature and corpus-based semantic-level features, of which the combination can mine more information of interest and simplify the labor-intensive process compared with universal fields.
 \item Experiment results show that compared with the existing methods, MedSim can evaluate similarity more effectively. This reveals that on the analytics and assessments of semantic network, domain specific features, structural and textual information are important.
\end{itemize}

The rest of this paper is organized as follows: Section 2 presents the process of dataset. Section 3 proposes our semantic similarity method MedSim based on bio-medical KGs. Section 4 reports the evaluation experiments and explains the evaluation results. Conclusions and future work are outlined in Section 5.

\section{Data Processing}
\subsection{Preparation of Data Source}
Semantic similarity measures relied on single data source provide only partial information about a subset of interest and the computed results show various degree of incompleteness. To address this problem, MedSim integrates the following well-established and widely used multi-category data sources.
\subsubsection{DrugBank} DrugBank ~\cite{ref_article25} is a comprehensive bioinformatics and cheminformatics KG that combines detailed drug entities with drug information. It contains 10,513 drug entities including 1,739 approved small molecule drugs, 873 approved biotech (protein/peptide) drugs, 105 nutraceuticals and over 5,029 experimental drugs. Each drug entity contains more than 200 properties, such as chemical structure, prescription, pharmacology, pharmacoeconomics, spectra, etc. 
\subsubsection{SIDER} SIDER~\cite{ref_article26}  is a side effect database of information on marketed medicines and their recorded adverse drug reactions, including 1430 drugs and 5868 side effects. The relationships between antibiotics and the corresponding side effects will be extracted to calculate the side effect based similarity in MedSim. 
\subsubsection{NDF-RT} National Drug File – Reference Terminology (NDF-RT) ~\cite{ref_article27} combines the hirerachical drug classification with multiple drug characteristics including physiologic effect, mechanism of action, pharmacokinetics, etc. We extract mechanism of the essential pharmacologic properties of medications (physiologic effect and mechanism of action) from NDF-RT.
\subsubsection{PubMed} PubMed ~\cite{ref_article28} is a bio-medical search engine accessing more than 27 million citations for biomedical literature from MEDLINE, life science journals, and online books. We crawl more than 500,000 papers about medicine via the PubMed API to help establishing the semantic features of MedSim. 
\subsection{Antibiotic Pairs Labeling}
To verify the effectiveness of MedSim, we conduct experiments on 52 most commonly used antibiotics of 10 categories in hospital. With the combination of these antibiotics, 1326 pairs are generated. 528 randomly selected pairs cover nearly 40\% of the total. Referring to ~\cite{ref_article29,ref_article30}, doctors, from the perspective of clinical application, score the similarity between two antibiotics, which ranges in [0, 1], according to both antibacterial spectrum and efficacies of medicine. 0 indicates that there is no similarity between two antibiotics, while 1 implies that the two antibiotics are extremely similar. The adverse reactions, side effects, patient’s past history and other factors are left aside in this stage. To make antibiotic pairs labeling more accurate, each pair is labeled by at least 3 doctors and the average is taken as final result. The Pearson coefficient between the scores issued by each doctor and the average score ranges from 82.7\% to 86.4\% while Spearman coefficient ranges from 79.2\% to 88.8\%, both proving the reliability of doctors' assessment. Scores about the antibiotic similarity were uploaded to Github.The labeled antibiotic pairs are divided into training set and test set, which will be used in our regression prediction model in Section 4.

\section{Methodology}
MedSim is a medicine similarity metric predicted by the random forest regression model learned from the following features: medicine-specific features (side effect, target protein, mechanism of action, physiological effect), structure feature of concept taxonomy (hierarchy embedding-based feature), and semantic-level features (KG-based semantic textual similarity and word embedding-based feature).

\subsection{Medicine-specific Features (MF)}
Domain-specific KGs and multi-category corpus are adopted to mine medical-specific features, so as to address the incompleteness of drug attributes from single data source. Side effect, target protein, mechanism of action, and physiological effect are utilized to explore the medicine-specific features which can simplify the semantic representation of medicines. 

\begin{figure}
\centering
\includegraphics[width=0.65\textwidth]{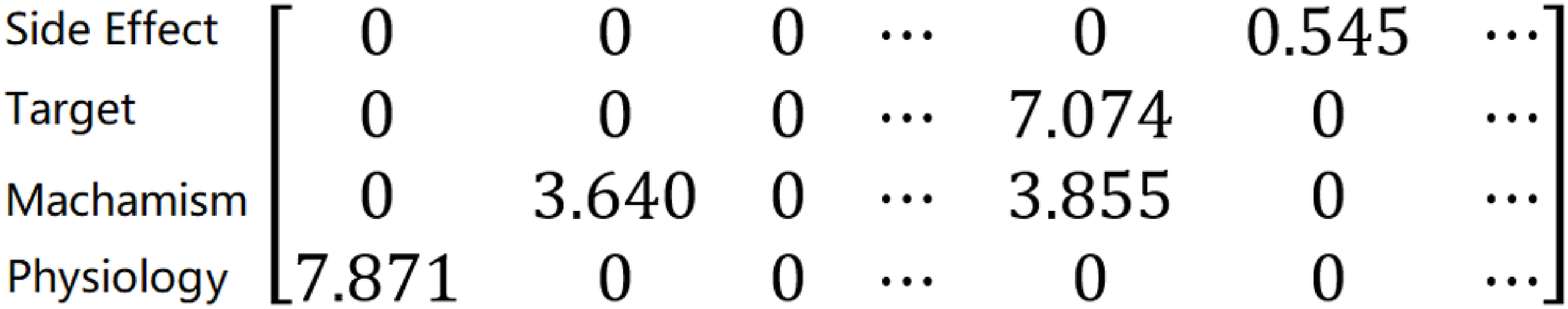}
\caption{Part of multi-dimensional weighted feature vector of  nitrofurantoin.} \label{Fig1}
\end{figure}

Instead of simply flattening all properties into one vector, the weighted property vectors of different features from multiple data sources are generated to interpret the characteristic of drugs. For a drug, its multi-dimensional weighted feature vector is erected by stacking weighted vectors of all medicine-specific features. Each row of vector represents one category of characteristics, in which the values demonstrate the weights of specific properties. Fig.~\ref{Fig1} shows a snapshot of multi-dimensional weighted feature vector of an antibiotic: nitrofurantoin.

\subsubsection{Side Effect based Similarity}
For a drug d, its side effects can be obtained from SIDER database. In the paper, we want to find out side effects related to some drug, as well as those specific to this certain drug, hence improve the discrimination. Inverse Document Frequency (IDF) can work well in alleviating the impact of high frequency terms and pay more attention to rare ones:
\begin{equation}
IDF(s,Drugs)=\frac{log(|Drugs|+1)}{DF(s,Drugs)+1}
\end{equation}

Where $⁡Drugs$ is the set of all drugs, $s$ ⁡is a side effect, and $DF(s, Drugs)$ is the number of drugs with the side effect
sider $s$ . The weighted side effect vector of a drug $d$ is $sider(d)$, consisting of side effects extracted from SIDER. The value of element $s$ of $sider(d)$, denoted $sider(d)[t]$, is $IDF(s, Drugs)$ if it is one of the side effects of drug d, otherwise it is $0$. The side effect-based similarity of two drugs $d_{1}$ , $d_{2}$ is the cosine distance of the vectors $sider(d_{1})$ and $sider(d_{2})$.

\subsubsection{Target based Similarity} The information about proteins targeted by a drug $d$ is collected from DrugBank. The target-based similarity of two drugs $d_{1}$, $d_{2}$ is defined as the cosine similarity of IDF-weighted target protein vectors of two drugs, which are calculated like the IDF-weighted side effect vector.

\subsubsection{Mechanism based Similarity} We collect all the mechanisms of a drug from NDF-RT.  The mechanism-based similarity of two drugs is calculated by the cosine distance of IDF-weighted mechanism vectors of two drugs as mentioned in the previous paragraph.
\subsubsection{Physiological Effect based Similarity} The IDF-weighted physiological effect vectors of drugs are also established from NDF-RT, and physiological effect-based similarity measure is the same as the mechanism-based similarity. 

\subsection{Hierarchy Embedding-based Feature (HF)}
To fully take advantage of taxonomy hierarchy information, DeepWalk ~\cite{ref_article31} is applied to learn the hierarchy embedding from the taxonomy hierarchy in DrugBank. All concepts in taxonomy hierarchy are actually arranged as a concept graph, where nodes represent concepts and edges indicate hierarchical relations. 

Unlike some structure-based methods which focus on concrete properties of KG, e.g. neighbors, class hierarchies and node degrees, hierarchy embedding based feature is able to map a total knowledge graph into a low dimension vector space while preserving certain properties of the original graph. Due to the limited depth of taxonomy hierarchy, the training paths in DeepWalk can cover both complete leaf to root path and neighbor nodes, which can fully utilize structural information. The hierarchy embedding-based similarity is calculated by the cosine distance between two drug vectors.

\subsection{Semantic-level Features (SF)}
We adopt two types of semantic-level features: the KG-based semantic textual similarity feature is used to process the entity related textual information, while the word embedding-based similarity feature learned from the context of drugs is used to cluster similar entities in vector space.
\subsubsection{KG-based Semantic Textual Similarity (KSTS)} Traditional IC-based semantic similarity metrics require a large domain corpus and cost-intensive labor to remove redundant data. KGs have already mined topic-related knowledge from textual corpus, which has prepared a high-quality domain corpus. The entity description or other textual information about the concepts in KG usually implies the nature of the concepts. The greater is the similarity among the concepts, the greater is the similarity of the words in their entity description.

The bio-medical proper nouns in the entity description of universal KG tend to have few information, which cannot improve the performance of the semantic textual similarity measures ~\cite{ref_article32}. In this context, BM25 algorithm ~\cite{ref_article33} is applied to compute the textual similarity based on entity description by converting the ranking score into the similarity score that ranges between 0 and 1.

Given a description of a drug $d_{1}$ containing the keywords $q_{1}$, $q_{2}$ , ... , $q_{n}$ , the BM25 score of a description $D$ of another drug $d_{2}$, is

\begin{equation}
score(d_{1},d_{2})= \sum_{i=1}^{n}IDF(q_{i})\cdot \frac{ f(q_{i},D)\cdot (k_{1} + 1)}{f(q_{i},D)+k_{1}\cdot(1-b+b\cdot \frac{|D|}{avgdl})}
\end{equation}

where $IDF(q_{i})$ is the IDF weight of the keywords $q_{i}$, $f(q_{i} , D)$ is the occurrence frequency in the description $D$, $|D|$ is the number of words in $D$, and $avgdl$ is the average number of words of all entity descriptions drawn from DrugBank. Usually, the free parameters $k_{1}\in [1.2, 2.0]$ and $b =
0.75$. Here, we set $k_{1} = 2$ and $b = 0.75$. To normalize the BM25 ranking score, the KG-based Semantic Textual Similarity is definded below, where $Drugs$ is the set of all drugs:
\begin{equation}
KSTS(d_{1}, d_{2})=\frac{score(d_{1}, d_{2})-min\{score(x, y)|x,y \in Drugs \}}{max\{score(x, y)|x,y \in Drugs \}-min\{score(x, y)|x,y \in Drugs \}}
\end{equation}

The entity description included in DrugBank is few in words but large in numbers. Thereby, BM25 can quickly
measure the textual semantic similarity between a drug and the rest.

\subsubsection{Textual Embedding-based Similarity} Word2vec is applied to train the textual embedding vector of PubMed 500,000 indexed papers and medical corpus (e.g. DrugBank and DailyMed). We use the skip-gram model rather than CBOW according to the pre-experimental results. Since word2vec can predict concept relatedness by simple algebraic operations in vector space, the word embedding-based similarity is calculated by the cosine similarity between vectors
of corresponding drugs.

\subsection{Random Forest Regression Model}
Based on the aforementioned features, random forest regression model is applied to measure the medicines semantic similarity. Random forest is an effective ensemble learning algorithm for regression task. In this paper, random forest is used to predict the similarity of an antibiotic pair (a scalar dependent variable $y$) learned from the selected features of samples (explanatory variables $X$). 

Ten-fold cross validation is applied to train the model. The training and test datasets are from the labeled antibiotic pairs mentioned in Section 3.2, except some pairs whose labeled scores substantially differ. The Root Mean Square Error (RMSE), Mean Absolute Error
(MAE), Pearson correlation coefficient and Spearman rank correlation coefficient are adopted to evaluate the model. To select the best regression model, we make a detailed comparison of the random forest regression model and
other common methods such as linear regression, logistic regression, polynomial regression, tree regression, etc. 

As shown in Table~\ref{tab1}, the model that we employed has the
lowest RMSE and MAE, which indicates that it outperforms other models in precision and stability. The Pearson  and Spearman coefficient specify that the similarity measured by the random forest has a strong correlation with the results scored by doctors. Compared to other regression models, random forest is not very sensitive to missing data, which alleviates the impact from the incompleteness of drug attributes. Besides, the randomized sampling before bagging and the application of averaging can avoid overfitting and further improve the generalization ability.
\begin{table}
\centering
\setlength{\tabcolsep}{2mm}{
\caption{Performance comparison of regression models.}\label{tab1}
\begin{tabular}{ |l | l | l | l | l |}
\hline
Regression model &  Person & Spearman & RMSE & MAE\\
\hline
Logistic Regression &  0.273 & 0.239 & 2.391 & 1.641 \\
Linear Regression &  0.306 & 0.232 & 0.186 & 0.137 \\
Polynomial Regression &  0.431 & 0.442 & 0.178 & 0.137 \\
Support Vector Regression &  0.456 & 0.435 & 0.175 & 0.124 \\
Adaboost Regression &  0.465 & 0.435 & 0.172 & 0.129 \\
Bagging Regression &  0.520 & 0.474 & 0.164 & 0.123 \\
Tree Regression &  0.563 & 0.510 & 0.161 & 0.116 \\
\textbf{Random Forest Regression} &  \textbf{0.584} & \textbf{0.518} & \textbf{0.156} & \textbf{0.116} \\
\hline
\end{tabular}}
\end{table}

\section{Evaluation}
To assess the performance of our model, we conduct two types of experiments. We first compare it with the state-of-art semantic similarity methods to prove whether MedSim outperforms others. Furthermore, we measure the prediction ability of individual features and
different feature combinations to exploit the effect of the model performance. Spearman and Pearson correlations coefficients are widely used to evaluate semantic similarity measures. In this section, both coefficients are adopted to evaluate the correlation between doctor assessment and experiment results, while Z-significance test between MedSim and baselines is used to evaluate whether MedSim statistically outperforms other methods using two-sided test and 0.05 statistical significance.

\subsection{Comparison with State-of-art Similarity 
Metrics}
We compare MedSim with four state-of-art algorithms, including GADES ~\cite{ref_article22}, Res ~\cite{ref_article12}, Wpath ~\cite{ref_article19} and Hybrids ~\cite{ref_article24}.
The GADES is a structure-based measure, while Res is an
information content based measure. Wpath considers both
path and IC information. Based on Wpath,
the method Hybrids takes medical properties into account
to calculate the drug similarity.

As shown in Table~\ref{tab2}, different semantic similarity method
has different level of correlation between doctor’s judgment
and MedSim outperforms the others.

\begin{table}
\centering
\setlength{\tabcolsep}{2mm}{

\caption{Comparison between semantic method.}\label{tab2}
\begin{tabular}{|l|l|l|l|l|}
\hline
\multirow{2}*{Method} & \multirow{2}*{Pearson} & \multirow{2}*{Spearman} & \multicolumn{2}{|c|}{Z significance test}\\
\cline{4-5}
& & & Z statistic& p-value\\

\hline
GADES &  0.251 & 0.203 & 2.881 & 0.051\\
Res &  0.211 & 0.223 & 0.273 & 0.000\\
Wpath &  0.251 & 0.205 & 0.805 & 0.000\\
Hybrids &  0.256 & 0.278 & 0.995 & 0.000\\
\textbf{MedSim} &  \textbf{0.586} & \textbf{0.523} & N/A & N/A\\
\hline
\end{tabular}}
\end{table}

GADES has the lowest Spearman correlation,
probably due to the structure of bio-medical KGs. It is common sense that in KGs, the upper-level concepts in a taxonomy are supposed to be more general hence have more
entities. However, it may be different in bio-medical KGs,
where the entity number of lower-level concepts would be
larger than that of upper-level concepts. For example, according to concept tree in the latest released version (version 5.0.6) of DrugBank, the level of Tetracyclines is upper
than that of Aminobenzene sulfonamides. However, the entity number of the former one is 3246, which is far less than
that of the latter one, which has 235515 entities. Thereby,
structure based approach like GADES cannot work well in
the bio-medical KG-based similarity measures. Res also has the lowest Pearson correlation, which also implies the limited effect of IC-based measures in computing
medicine semantic similarity. 

Wpath shows a slightly improvement over GADES and
Res by adopting both structure information and semantic
information of KGs. When we set Wpath’s free parameter k=0.85, Wpath can achieve its own highest correlation score.

The Hybrids method takes all aforementioned medicine-specific features into account to measure the semantic similarity. Its highest score among all baselines indicates the
significance of the medicine-specific features.

Both Pearson and Spearman coefficient of MedSim are
over 0.5, indicating that the prediction of our model has a
high correlation with doctors’ judgment. Compared
with other methods, MedSim can more effectively evaluate similarity. The results of the z-test also
show that MedSim has a statistically significant improvement over baselines, since in each baseline z statistics are
larger than 1.96 and p-values are below the significance
level of 0.05. Experiment results also reveal that on the analytics and assessments of KG semantic/structure information, domain specific features need
to be considered simultaneously.

\subsection{Feature Selection Comparison}
The prediction ability of each feature and feature combinations is measured in this section (Table~\ref{tab3}).
\begin{table}
\centering
\setlength{\tabcolsep}{2mm}{
\caption{Comparison of feature performance.}\label{tab3}
\begin{tabular}{ |l | l | l | l | l |}
\hline
 & Feature & Person & Spearman\\
\hline
\multirow{3}*{Single Feature} & MF & 0.407 & 0.389\\
                                     & HF & 0.159 & 0.150 \\
                                     & SF & 0.339 & 0.258 \\
\hline
\multirow{3}*{Multiple Feature} & MF and HF & 0.551 & 0.489\\
                                     & MF and SF & 0.570 & 0.515 \\
                                     & \textbf{MF, HF and SF} & \textbf{0.585} & \textbf{0.523} \\
\hline
\end{tabular}}
\end{table}

For the single feature, the coefficient scores indicate that medicine-specific features yield a good performance without cooperating with other features .

The combinations of MF and HF and the combinations of MF and SF generally show much better performance than using these features separately, increasing the coefficients by at least 10\%. The best performance is obtained by the combination of all features, indicating that the proper combination of features can mine more information and improve the prediction performance.

There are four types of medicine-specific features adopted in our study, among which, the physiological effect based similarity with 21.3\% Pearson coefficient outperforms other features. The using of each individual feature cannot yield a satisfactory result. Especially, the removal of the physiological effect based similarity weaken the prediction performance of model by decreasing the Pearson coefficient by 15.5\%. Through various pre-experimental results, we believe that the current combination of
medicine-specific features is the one that is much helpful
in the semantic similarity calculation of biomedicine.

\subsection{Case Study}
To study the medicine substitution, we employ MedSim to predict the similarity scores between
cefoperazone and other 51 antibiotics. All pairs containing
cefoperazone are excluded from training set and considered as test set.

For the antibiotic cefoperazone, Table~\ref{tab4} presents its similar antibiotics whose similarity score is over 0.85. Refer to ~\cite{ref_article34,ref_url1}, the experiment results show that two antibiotics
whose similarity scores over 0.85 can be replaced by each
other under normal circumstances. We list the similar antibiotic names, provide the semantic similarity scores between antibiotics and cefoperazone evaluated by MedSim,
and present the cases where they can replace each other.

\begin{table}
\centering
\setlength{\tabcolsep}{2mm}{
\caption{Parts of antibiotics similar can replace cefoperazone}\label{tab4}
\begin{tabular}{ |l | l | p{0.65\columnwidth}|}
\hline
Similar Antibiotic &  Score & Cause where the antibiotic can replace cefoperazone\\
\hline
Cefoxitin &  0.865 &  Respiratory tract infections; Urinary tract infections; Peritonitis; Septicemia; Gynecological infections; Bone, joint, and soft tissue infections \\
\hline
Cefepime &  0.864 &  Respiratory tract infections; Urinary tract infections; Abdominal infections; Reproductive tract infections; Bone, joint, and soft tissue infections\\
\hline
Ceftriaxone &  0.860 &  Lower respiratory tract infection; Urinary tract infections; Complicated intra-abdominal infections; Infections in obstetrics and gynecology; Skin and soft tissue infections; Meningitis\\
\hline
Meropenem &  0.851 &  All infections of cefoperazone, but the has stronger efficacy and wider antibacterial
spectrums\\
\hline
\end{tabular}}
\end{table}

Take cefoperazone and ceftriaxone as an example. The
indication of cefoperazone is very close to ceftriaxone except disease caused by a few bacteria such as Pseudomonas aeruginosa. In the absence of susceptibility
testing, doctors can choose either of them to treat most of
Gram-negative bacteria infections, such as meningitis,
pneumonia and bronchitis. Once the inventory of either is
insufficient, our method can help doctors to find a most
similar one for replacement.

Another example is cefoperazone and cefpime, both of
which have good activity against Pseudomonas aeruginosa.
However, the combined application of them
cannot enhance the efficacy and is considered as
drug abuse. Quantifying the similarity of antibiotics, such
as listing antibiotics which have similar spectrum of bacterial susceptibility, may help improve public
understanding that sometimes antibiotics combination
should be avoided. Thus, medicine semantic similarity
measure can ease the increasingly serious problem of antibiotic abuse.

Though meropenem can replace cefoperazone clinically,
the semantic similarity score
is slightly over 0.85. The reason is that meropenem's indications far exceed cefoperazone. In other words, cefoperazone can not replace meropenem completely. The
meropenem will be applied to replace cefoperazone only when the infective bacteria exceeded the antibacterial
spectrum of cefoperazone or cefoperazone is ineffective.
Otherwise,  the replacement
of cefoperazone with meropenem is clinically an abuse of antibiotics
with higher antibacterial activity.

\section{Conclusion}
In this study, we propose MedSim, a novel semantic similarity method based on public well-established KGs and
large-scale drug corpus. MedSim fully utilizes not only 
the structural and textual features from the KG but
also medicine-specific features. MedSim produces statistically significant improvements over other methods. Examples of case study indicate that
calculating the medicine semantic similarity owns a prospect in therapeutic substitution and decreasing the problem of drug abuse. The proposed method is extensible, reproducible and
applicable to the KG-based similarity calculation in medical
field. Assuming that a drug can be located in both a medical search engine and a bio-medical KG, all the features used in MedSim can be immediately obtained and used to measure other types of medicine in addition to antibiotics. All features used in MedSim can be obtained from the public knowledge source and the labeled dataset is now freely accessible, thus, our method can be conveniently reproduced. In the future, we explore the performance of MedSim in other types of medicine, such as sedative once we get the labeled respective drug dataset. 
%
%
%
%

\end{document}